\documentstyle[aaai]{article}
\newcommand{\pdot}{\stackrel{.}{+}}
\newcommand{\ga}{\alpha} \newcommand{\gb}{\beta}

 \newcommand{\cl}{{\cal L}}

\newtheorem{proposition}{\bf Proposition}

\newtheorem{observation}{\bf Observation}

\newtheorem{definition}{\bf Definition}

\newcommand{\qed}{\vrule height5pt width3pt depth0pt}

\begin{document}

\title{Relevance Sensitive Non-Monotonic Inference on Belief Sequences}
\author{
Samir Chopra\\CUNY Graduate Center\\365 Fifth Avenue\\New York, NY
10016\\schopra@gc.cuny.edu\\
\And
Konstantinos Georgatos\\John Jay College of CUNY\\899 Tenth Avenue\\New York, NY 10019\\
kgeorgatos@jjay.cuny.edu\\
\And
Rohit Parikh\\CUNY Graduate Center\\365 Fifth Avenue\\New York, NY
10016\\ripbc@cunyvm.cuny.edu}
\date{January 15, 2000}

\maketitle
     
\begin{abstract} We present a method for relevance sensitive non-monotonic inference from
 belief sequences which incorporates insights pertaining to prioritized
 inference and relevance sensitive, inconsistency tolerant belief revision.
 Our model uses a finite, logically open sequence of propositional formulas
 as a representation for beliefs and defines a notion of inference from
 maxiconsistent subsets of formulas guided by two orderings: a temporal
 sequencing and an ordering based on relevance relations between the
 conclusion and formulas in the sequence.  The relevance relations are
 ternary (using context as a parameter) as opposed to standard binary
 axiomatizations.  The inference operation thus defined easily handles
 iterated revision by maintaining a revision history, blocks the derivation
 of inconsistent answers from a possibly inconsistent sequence and maintains
 the distinction between explicit and implicit beliefs.  In doing so, it
 provides a finitely presented formalism and a plausible model of
reasoning for automated agents.
\end{abstract}
\noindent

 \section{Introduction}\label{sec:introduction}

 Belief revision is the process of transforming a belief state $K$ upon
 receipt of new information $\ga$.  There are two fundamental approaches to
 this problem.  In the {\it logic-constrained} or {\it horizontal} approach
 \cite{gr95}, the belief representation is a {\it theory}\, $K$ and given a
 new proposition $\ga$, \cite{agm85} propose postulates for $K*\ga$, the
 theory revised with $\ga$.  In this approach the belief state is itself
 sophisticated, and constructing the updated belief state requires work.
The usual constructions for AGM revisions using selection
functions and
epistemic entrenchments often fail to provide an adequate account of {\it
iterated revision}.  AGM-like postulates do not specify how we {\it came}
to believe $K$ and after revision, it is assumed that $K*\ga$ is a generic
theory.  But these postulates ignore the fact that $\ga$ was our {\it last}
information.

 The {\it vertical} approach, in contrast, uses trivial (and repeatable)
 operations of revision and expansion on finite, logically open, belief
 representations, but utilizes a {\it sophisticated} notion of non-monotonic
 inference, see, e.g. \cite{do79}, \cite{bre91}.

 We suggest, in conformance with the vertical approach, that $K*\ga$ be
 taken to be the {\it belief sequence} $K;{\ga}$, i.e., a {\it finite},
 logically open, sequence of propositions with $\ga$ being the most recent.
 This suggestion that the most perspicuous way to represent our beliefs is a
 finite, logically open, set of sentences, i.e., a {\it belief base} has been
 made (amongst others) by \cite{han92}, \cite{neb92}; the notion that a
 sequence of formulas captures the importance of temporal ordering and of
 maintaining a revision history is noted by \cite {ry91} and \cite{leh95}.

 Since updating becomes simple under this approach, the notion of inference
 must be correspondingly more sophisticated.  We describe a method for
 non-monotonic inference from belief sequences which does this, but departs
 significantly from previous approaches in one respect.  It makes heavy use
 of {\it relevance relations} amongst formulas in a belief base.

 This method incorporates the insights in earlier proposals made by
 \cite{kg96}, \cite{par96} and \cite{cp99}. \cite{kg96} uses the linear
 order of a belief sequence as a prioritization to generate a variety of
 inference relations, shows that these schemes are non-monotonic
and, therefore, induce a method for belief revision.
 \cite{par96} shows that if we have a theory referring to two or more
 disjoint subjects, then our language can be partitioned into corresponding
 sub-languages, and it is suggested that new information about {\em one}\,
 of them should not affect any other.  This ensures a relevance or context
 sensitive, {\it localized} notion of belief revision and serves as one way
 of capturing a more general notion of relatedness amongst propositions in a
 belief base (as studied by \cite{was99}). \cite{cp99}
consider sets of theories
called $B$-structures, which are individually consistent, but can be {\em
jointly} inconsistent, to capture the intuition that real agents often
reason with an inconsistent, yet usable, set of beliefs which is divided
into individually consistent compartments.

 Our (current) method of inference blocks the derivation of explicitly
 inconsistent beliefs from a possibly inconsistent belief sequence by using
 a notion of inference from {\it maxiconsistent subsets of relevant
 formulas}.  Choosing maxiconsistent subsequences in order to avoid
 inconsistency was used in \cite{kg96}, while relevance is determined, as in
 \cite{par96} and \cite{cp99} by a specialized notion of language overlap or
 by other context determined features.  The formula whose inference from the
 sequence is to be determined imposes a prioritization on the formulas
 present in the sequence by virtue of its relevance relations with them
 (thus reorganizing the temporal ordering present in the sequence).

 Thus we do not treat a belief sequence as a set but rather as a
 linear order much like an entrenchment (\cite{gm88}, \cite{kg97})
 except that {\it two} orderings, level of relevance and temporal order
 govern the sequence. The resulting procedure for inference serves as a
 generalization of the methods presented in \cite{kg96} and \cite{cp99}.  In
 this way, we hope to present a model for belief revision that is a
 plausible representation of real agents' reasoning.

 In the first section of the paper, we present preliminary definitions and
establish the notion of relatedness that we
 will work with.  In the second section we define our notion of inference
 and examine its properties.\vspace{.05in}

 \noindent {\bf Notation:} In the following, ${\cal L}$ is a finite
 propositional language with the usual logical connectives ($\neg, \vee,
 \wedge, \rightarrow, \leftrightarrow$).  The constants {\it true, false}\,
 are in ${\cal L}$. Greek letters $\alpha, \beta, \gamma$ denote arbitrary
 formulas while Roman lower case letters $p, q, r$ denote propositional
 atoms. $\alpha \Leftrightarrow \beta$ means that $\ga \leftrightarrow \gb$
 is a tautology. $Cn$ will denote the usual classical consequence relation.
 We reserve the letters $\sigma, \tau$ for belief sequences.

 \section{Belief Sequences}

We begin with a definition of a belief sequence:
 \begin{definition} A belief sequence is a sequence of formulas under a
 temporal ordering, i.e., a sequence of formulas, $\sigma =
 \beta_1\ldots\beta_n$ where for any pair of beliefs $\beta_i, \beta_j$ if
 $i < j$, $\beta_j$ is more recent than $\beta_i$.  Given two sequences,
 $\sigma_1, \sigma_2$ we say that $\sigma_1 \sqsubseteq \sigma_2$ if
 $\sigma_2$ is obtained from $\sigma_1$ by the concatenation of zero or more
 formulas; $\sigma_1$ will be referred to as an initial segment of
 $\sigma_2$\end{definition}
Under
 a temporal ordering the most recent formulas occur at the tail of the
 sequence.  We assume that each formula $\alpha$ in the sequence is
 expressed in its {\it smallest language} $L_{\alpha}$ as defined below.
Note that the linear temporal order can be replaced with some other linear
order expressing prioritization. For example, one could order the
propositions on the basis of the trustworthiness of their source.

 We now present a relation of {\it relatedness} amongst formulas in a
 sequence (originally proposed and used in \cite{par96}, \cite{cp99} for
 localized belief revision) as a preliminary to its modification for use in
 this study.  First, a distinction between different languages that a
 formula can be expressed in:
 \begin{definition} The language $L(\alpha)$ is the set of propositional
 variables which actually occur in a formula $\alpha$; the language
$L_\ga$
 of $\ga$ is the {\em smallest} set of propositional variables which can be
 used to express $\beta$, a formula logically equivalent to $\ga$.
 \end{definition}
 So, if $\ga=p_1\wedge (p_2 \vee \neg p_2)$ then $\beta$ is $p_1$ and
 $L_\ga=\{p_1\}$ while $L(\alpha) = \{p_1, p_2\}$.  ($L_\alpha$ is unique,
 cf.  Lemma LS1 in \cite{par96}).  $L_{\alpha}$ has logically
 attractive properties, e.g. if $\alpha \Leftrightarrow \beta$, then
 $L_{\alpha} = L_{\beta}$.  Hence we shall work exclusively with this
 notion.
 \begin{definition} $\alpha, \beta$ are related by syntactic language
 overlap (${\cal R}_s (\alpha, \beta))$ if $L(\alpha) \cap L(\beta) \neq
 \emptyset$. $\alpha, \beta$ are related by logical language overlap (${\cal
 R}_l (\alpha, \beta)) $if $L_{\alpha} \cap L_{\beta} \neq \emptyset$.
 \end{definition}
It is easily seen that
${\cal R}_l (\alpha, \beta))$ implies ${\cal R}_s (\alpha, \beta))$.

 In  earlier work on relatedness, Epstein \cite{eps95} imposes the following
    `plausibility' requirements
 on any relatedness relation $\cal R$$(\alpha,\beta)$ ($\alpha$ is related
 to $\beta$):\\ \noindent {\bf R1} $\cal R$$(\alpha,\beta)$ iff
 $\cal R$ $(\neg \alpha,\beta)$.\\ \noindent{\bf R2} $\cal R$$(\alpha,\beta
 \wedge \gamma)$ iff $\cal R$ $(\alpha, \beta\rightarrow \gamma)$.\\
 \noindent {\bf R3} $\cal R$$(\alpha,\beta)$ iff $\cal R$$(\beta,\alpha)$.\\
 \noindent {\bf R4} $\cal R$$(\alpha,\alpha)$.\\ \noindent {\bf R5} $\cal R$
 $(\alpha, \beta\rightarrow \gamma)$ iff $\cal R$ $(\alpha,\beta)$ or $\cal
 R$$(\alpha,\gamma)$.\\

\noindent  Rodrigues \cite{rod97} has shown that the relation (${\cal R}_s (\alpha, \beta))$,
i.e., syntactic language overlap, is the smallest relation satisfying
    Epstein's conditions.

    \noindent However, we might also consider a condition not considered by
    Epstein: \\{\bf R6} if $\cal R$$(\alpha,\beta)$ and $\beta
    \Leftrightarrow {\beta}'$ then $\cal R$$(\alpha,{\beta}')$.

 \begin{observation}${\cal R}_s$ does not satisfy {\bf
 R6} whereas the relatedness relation ${\cal R}_l$ does satisfy {\bf R6} as
 well as conditions {\bf R1, R3-4} (but not the conditions {\bf R2, R5}).\end{observation}

Since condition {\bf R6} is very natural, we wonder if conditions {\bf R2,
R5} were adopted out of a feeling that they are compatible with {\bf R6}.
Indeed, in general, though not always, we do have $L_{\beta \wedge \gamma}
= L_{\beta \rightarrow \gamma} = L_{\beta} \cup L_{\gamma}$.  In such a
 case it will be the case that $\alpha$ is relevant to the composite formula
 iff it is relevant to at least one part.  Note that our $R_l$ does satisfy {\em half}\, of {\bf R5}:\\
{\bf
R5a:} If  ${\cal R}_l(\alpha, \beta\rightarrow \gamma)$ then ${\cal R}_l
 (\alpha,\beta)$ or ${\cal R}_l(\alpha,\gamma)$.

 For an actual example, notice that if we let $\beta = p, \gamma = \neg p$,
 then $\beta \rightarrow \gamma$ is equivalent to $\neg p$ and of course
 relevant to $p$.  However, $\beta \wedge \gamma$ is $p \wedge \neg p$, a
 downright contradiction and not relevant (in our opinion) to {\em
 anything}.  This fact casts some doubt on the intuition behind {\bf R2}.
 Indeed it turns out that {\bf R2,R5} are incompatible with the natural
 requirement {\bf R6}.\vspace{.1in}

 With the discussion above as a guide, we now
 develop a {\it context-sensitive} measure for {\it relevance} amongst
 formulas in a belief sequence.
 \begin{definition}Two formulas $\gb_1,\gb_2\in \cl$ are (logically)
 {\em disjoint} iff $L_\ga\cap L_\gb=\emptyset$. \end{definition}
 \begin{definition}\begin{enumerate}
\item A pair of formulas, $\alpha, \beta$ are directly
 relevant if they are not logically disjoint, i.e., if $L_\alpha \cap
L_\beta
 \neq \emptyset$.
\item Given a belief sequence $\sigma$, a pair of formulas
 $\alpha, \beta$ are $k$-relevant wrt $\sigma$ if $\exists \chi_1, \chi_2,
 \ldots \chi_k \in \sigma$ such that:\\ \noindent i) $\alpha, \chi_1$ are
 directly relevant\\ \noindent ii) $\chi_i, \chi_{i+1}$ are directly
 relevant for $i = 1, \ldots k-1$\\ \noindent iii) $\chi_k, \beta$ are
 directly relevant.\\ \noindent We write ${\cal R}_k(\alpha, \beta,
 \sigma)$ to indicate that $\alpha, \beta$ are $k$-relevant w.r.t
$\sigma$. If $k=0$ above, the
formulas are directly
 relevant.
\item A pair of formulas are irrelevant if they are not $k$-relevant
 for any $k$.

\item $rel(\alpha,\beta,\sigma)$ is the the lowest $k$ such that
 $\alpha,\beta$ are $k$-relevant wrt $\sigma$ (we let it be $\infty$ if
 $\alpha, \beta$ are irrelevant).
\end{enumerate}
 \end{definition}

 In the following observations, we omit the sequence $\sigma$ when
 clear from the context.
 \begin{observation} \begin{enumerate}
 \item If a pair of formulas are
 $k$-relevant, then, $\forall m > k$, they are $m$-relevant as well. \item
 Let $[\![\sigma]\!] = \{ \alpha \mid \alpha$ occurs in $\sigma \}$.  We say
 that $\sigma_1 \subseteq^{'} \sigma_2$ iff $[\![\sigma_1]\!] \subseteq
 [\![\sigma_2]\!]$.  If $\sigma_1 \subseteq^{'} \sigma_2$, then
 $rel(\alpha,\beta,\sigma_2) \leq rel(\alpha,\beta,\sigma_1)$.
 \item The
 relation ${\cal R}_k(\alpha, \beta, \sigma)$ is both symmetric and
 reflexive in the first two arguments but obviously not transitive.
 \item If
 $k = 0$ then the sequence $\sigma$ is irrelevant to the question of
 $k$-relevance of formulas $\alpha, \beta$; if $k \neq 0$, then $\sigma$ is
 a parameter in determining relevance between $\alpha, \beta$. \end{enumerate}
 \item ${\cal R}_k(\alpha, \beta, \sigma)$ depends only on
 $L_{\alpha},L_{\beta}$ and $[\![\sigma]\!]$ and not on
 ${\alpha},{\beta}$ and $\sigma$ themselves.
\end{observation}
 So if two formulas are relevant to each other at one level, then they are
 relevant at all weaker ({\em higher}\, $k$) levels.  The definition of
 relevance above explicitly brings in (as a third parameter) the sequence
 $\sigma$ which can form a bridge between formulas which do not have a
 direct overlap. Normally, relevance has been thought of as a binary
relation; our definition
renders it a ternary one.  Thus the sequence $\sigma$ can play the role of
 connecting up formulas which do not have any overlap in language but are
 connected {\em through}\, other beliefs.  A fact about Taj Mahal and one
 about India will be connected because of our (true) belief that the Taj
 Mahal is in India.  We can also have more distant - and less convincing -
 indirect connections.  E.g.
 we can think of beliefs linking the subject matter {\it
 European History}, to {\it European Music} and then to {\it Music} in
 general, to {\it Indian Music}.  But it is unlikely that we have beliefs
 {\em directly} linking European History to Indian Music and this level ($k$
 = 2) may be too weak (too high) to be useful in most considerations.

 The above definition extends the definition of relevance proposed in
 \cite{par96} and makes explicit the {\it contextual nature} of the
 relevance definition: two formulas $\alpha, \beta$ may have different
 degrees of relevance to each other in virtue of different belief sequences.
 A belief sequence defines a particular context for subject matters; pairs
 of formulas acquire different relationships to one another given differing
 contexts.  The more basic beliefs (i.e., elements of $\sigma$) that a
 person has, the more likely (s)he is to connect two apparently unconnected
 subjects.

 \section {Revision and Inference on Belief Sequences} Revision on belief
 sequences is easily achieved: we simply concatenate the new formula
 $\gamma$ to the sequence.  The sequence $\beta_1, \ldots, \beta_n$ becomes
 $\beta_1 \ldots \beta_n, \beta_{n+1}$ where $\beta_{n+1} =
 \gamma$.\vspace{.05in}

 \begin{definition}$\sigma \ast \gamma = \sigma; \gamma$ where $\sigma;
 \gamma$ represents the concatenation of $\gamma$ to a belief sequence
 $\sigma$.  \end{definition}

\noindent  Note that $[\![\sigma \ast \gamma]\!] = [\![\sigma]\!] \cup \{\gamma\}$.

 \noindent {\bf Example:} Consider the sequence $\sigma = [p, q, p \wedge q,
 \neg r]$.  If we receive the information $\beta = \neg p$, it is simply
 appended to the sequence to give us $\sigma \ast \beta = [p, q, p \wedge q,
 \neg r, \neg p]$. $[\![\sigma]\!]$ is now inconsistent, and we need a notion
 of inference which renders the agent's beliefs coherent.

 \subsection{Prioritized Inference} Suppose that we have a sequence of
 formulas which is our current belief base and we are asked about some
 formula $\gamma$, whether we believe it or not.  As we saw just above, the
 set $[\![\sigma]\!]$ may well be inconsistent, and hence to decide about
 the status of $\gamma$ we will need to pick some consistent
 subset of $[\![\sigma]\!]$
 which we bring into play.  The choice of formulas to be in such a subset
 will be governed by two considerations.  One is temporality (which we
 provisionally adopt) under which more recently received formulas have
 priority over older formulas.  The other is relevance according to which
 more relevant formulas have priority.  Clearly we need to decide which order
 counts more.  We have made the decision in this paper that relevance is
 more important than temporality but that between two formulas of equal
 relevance, the more recent formula has priority, We concede however, that
 the other procedure, to count temporality more, also has something to say
 in its favor.  Ordinary human reasoning, in our opinion is a pragmatic
 blend of the two techniques.

 We use the {\it maxiconsistent} approach to define {\it prioritized}
 inference on a belief sequence $\sigma$.  This method employs
 a consistent subset of $[\![\sigma]\!]$ obtained as follows.
 Consider a formula $\gamma$ in
 its smallest language $L_\gamma$.  We construct a maxiconsistent subset
 $\Gamma_{\langle\sigma,k,\gamma\rangle}$ (of $k$-relevant to $\gamma$
 formulas) of $[\![\sigma]\!]$.  The construction of this set is regulated
 by the ordering $\prec$ that $\gamma$ creates on $\sigma$, which arranges
 $\beta_1, \ldots, \beta_n$ into $\delta_1, \ldots, \delta_n$ as follows:

 \begin{definition} Given a formula $\gamma$, a sequence $\sigma$,
 $\beta, \beta' \in \sigma$, $\beta \prec \beta'$ if either \\ a)
$rel(\gamma,\beta, \sigma) < rel(\gamma,\beta', \sigma)$ (i.e., $\beta$ is
more relevant to $\gamma$ than $\beta'$) \\or
 b) $\beta, \beta'$ are equally relevant (i.e., 
$rel(\gamma,\beta, \sigma) = rel(\gamma,\beta', \sigma)$)
 but $\beta$ is
more recently
 received than $\beta'$ \end{definition}
 The $\delta_1, \ldots, \delta_n$ are the $\beta_1, \ldots, \beta_n$ under
 this order.  In the definition below $\Gamma$ is (short for) the set
 $\Gamma_{\langle\sigma,k,\gamma\rangle}$ referred to above. $k$ is some
 preselected level of relevance.
\begin{definition}$\Gamma^0 = \emptyset$,\\ $\Gamma^{i+1} = \left\{
\begin{array}{ll} \Gamma^i & \mbox{if $\Gamma^i \vdash \neg\delta_{i+1}$
or if $\neg {\cal R}_k(\delta_{i+1}, \gamma, \sigma)$} \\ \Gamma^i \cup \{
\delta_{i+1}\} &
 \mbox{otherwise}
\end{array}
\right.$\\
\noindent $\Gamma_{\langle\sigma,k,\gamma\rangle}=
\Gamma^n$.
\end{definition}
 We check formulas for addition to $\Gamma_{\langle\sigma,k,\gamma\rangle}$
 in order of their decreasing relevance to $\gamma$.  The lower the level of
 relevance allowed (i.e., the higher the value of $k$), the larger the part
 of $\sigma$ considered.  We now define the inference operation $\vdash_k$.

 \begin{definition} $\sigma \vdash_k \gamma$ iff
$\Gamma_{\langle\sigma,k,\gamma\rangle} \vdash
 \gamma$ \end{definition}
 Once $\Gamma_{\langle\sigma,k,\gamma\rangle}$ has been constructed, the
 inference operation defined above enables a query answering scheme for the
 agent with definite responses:

 \begin{quote} If $\Gamma_{\langle\sigma,k,\gamma\rangle} \vdash \gamma$,
 then answer `yes'.\\ If $\Gamma_{\langle\sigma,k,\gamma\rangle} \vdash
 \neg\gamma$ then answer `no'.\\ Otherwise, answer `no information'.
 \end{quote}

 Even if $[\![\sigma]\!]$ is inconsistent, the agent is
 able to give consistent answers to every individual query.
 \subsubsection{\bf Discussion}
 The notion of inference thus defined has some desirable features.  As an
 example, suppose our belief sequence is initiated by first being told $p$
 and then $\neg p \wedge \neg q$. $\neg p \wedge \neg q$ overrides $p$ and
 we will no longer answer `yes' to $p$.  However, if we are now told $p \vee
 q$, this new information overrides $\neg p \wedge \neg q$.  Thus, our
 maxiconsistent set $\Gamma_{<\sigma, 0, p>}$ is $\{p \vee q, p\}$ and the
 query $p?$ will now be answered in the affirmative.  This is plausible
 since the latest information decreases the reliability of $\neg p \wedge
 \neg q$ and the original information $p$ regains its original standing.
 Such accomodations are not easily made within traditional, AGM-based
 frameworks for belief revision.


 The maxiconsistent set $\Gamma_{<\sigma, k, p>}$ obtained depends on
 whether some new information came ``in several pieces" or as a single
 formula.  Receiving two pieces of information individually and together
 often has very different effects on an agent's epistemic state.  If we
 receive $\alpha$ and $\beta$ seperately, then a later information which
 undermines $\alpha$ need not undermine $\beta$ as well.  But if we received
 the {\em conjunction} $\alpha \wedge \beta$ then undermining one will
 undermine both.  As an example, consider the arguments against the AGM
 Axioms 7,8 in \cite{par96} where it is argued that revision by conjunctions
 is not the same as revising by conjuncts individually.

 {\bf Example:}  Suppose our current beliefs are $\neg p, \neg q$ and we
 receive the information that $p \vee q$.  If $p$ is much more believable
 than $q$ we might now decide that $p$ holds.  Suppose we next hear that
 $\neg p \vee q$.  Since this is consistent with our current state, we
 accept it, ending with a state generated by $p,q$.  However, if we had
 received the {\em conjunction}  $(p \vee q)\wedge (\neg p \vee q)$, this
 would be equivalent to receiving just $q$ and we would never have believed
 $p$ at all.  Note that this is also a situation where we found Epstein's
 postulate {\bf R2} to be implausible, for the language
 $L_{(p \vee q)\wedge (\neg p \vee q)}$ does not equal
 $L_{(p \vee q)} \cup  L_{(\neg p \vee q)}$ in this case.

 For a concrete example let $p$ be ``The stove is smoking" and $q$ be ``The
 house is on fire".  If I am told that either the stove is smoking or that
 the house is on fire, I will choose to believe that the stove is smoking.
 If I later find that either the stove is not smoking or the house is on
 fire, since this belief is consistent, I must now add it, and conclude that
 the stove is smoking {\em and} the house is on fire!  But if I had been
 told the conjunction, which is equivalent to ``The house is on fire", I
 would never have thought about the stove at all.

 As a final point, consider the sequence $\sigma_2 = [p \wedge q, r \wedge
 \neg q]$.  Both $p$ and $r$ are derivable although their respective
 derivation is based on {\it incompatible} information.  This is due to the
 fact that the formulas which are preferred in each case depend on the
 query.  This is plausible however; agents often think about unrelated
 pieces of information in isolation from each other.  Here $p$ and $r$ are
 not directly related and therefore may be thought about by the agent
 separately.
 \subsubsection {\bf Prioritization of Directly relevant formulas} Further
 depth could be introduced into the method provided above by introducing a
 prioritization amongst formulas based on the {\it amount} by which formulas
 $\beta$ of $\sigma$ overlap with the query formula $\gamma$.  Thus such a
 prioritization $\preceq_{\gamma}$ might be defined as follows: let $\beta_1
 \preceq_{\gamma} \beta_2$ if $| L_{\beta_1} \cap L_{\gamma}| > |
 L_{\beta_2} \cap L_{\gamma}|$.  Under this prioritization scheme, we would
 use the size of overlap of languages as a measure.

 Such a measure however, is open to objections that propositions that share
 fewer symbols might actually be more relevant than those that share more
 symbols {\it depending upon the symbols shared}.  As an example, consider
 our intuition that if two formulas both mention {\it aardvarks}, they are
 (most likely) more relevant to each other than if they mentioned {\it
 cats}.  Or, the fact that the authors of this paper are logicians is a
 closer relatedness relation than the fact that they are all human beings.
 The ordering $\beta_1 \preceq_{\gamma} \beta_2$
 can be {\em further}\, refined then, by the following heuristic: if symbols
 shared by formulas occur frequently in $\sigma$, then there is a
 smaller likelihood that the formulas are relevant to each other whereas if
 they share symbols that occur with less frequency in the sequence $\sigma$.
 then the relvance is greater.
 This
 provides for a notion of {\it degree of relevance} amongst directly (and
 only for directly) relevant formulas.  However, a fuller discussion would
 take us too far afield and we postpone such refinements to a more extended
treatment.
 \subsubsection{\bf Answer sets and Consequence relations}
 We now define an attendant notion of a consequence relation $C_k(\sigma)$
 at a given level of relevance, $k$:
 \begin{definition}$C_{k}(\sigma) = \{\gamma | \sigma \vdash_k
 \gamma\}$\end{definition}
\begin{definition} $C(\sigma) = \{\gamma |
\exists k,  \sigma \vdash_k \gamma\} = \bigcup C_k(\sigma)$\end{definition}
 In view of the following propositions, if we were not worried about
 computational costs then $C(\sigma)$
 would be the only notion which would interest us as $C_k(\sigma)  \subseteq
 C_{k+1}(\sigma)$ and so we would only stop at a smaller $k$ to conserve
 resources.
 \begin{proposition} The inference
 procedure defined above is {\it monotonic} in $k$, the degree of relevance,
 i.e., $\sigma \vdash_k \gamma \Rightarrow \sigma \vdash_{k+1} \gamma$.
 \end{proposition}
 The above follows immediately from the fact that if $k < k'$ then
 $\Gamma_{<\sigma, k, p>} \subseteq \Gamma_{<\sigma, k', p>}$ It will also
 be useful to remember that if two formulas $\alpha,\beta$ have the same
 language, $L_\alpha = L_\beta$, then $\Gamma_{<\sigma, k, \alpha>} =
 \Gamma_{<\sigma, k, \beta>}$. \vspace{.05in}

 \noindent {\bf Remark:} The inference procedure is of course {\it
 non-monotonic} in expansions of a belief sequence, i.e., if $\sigma
 \sqsubseteq \sigma^{'}$ and $\sigma \vdash_k \gamma$, it is not necessarily
 the case that $\sigma^{'} \vdash_k \gamma$.  For example $p$ is derivable
 from the sequence $p$, but not from, say $p; \neg(p\vee q)$.
 But if we revise by formulas that are `irrelevant'
 to the query $\gamma$ in question then
 they will have no effect on the derivation of $\gamma$ and
 $\gamma$ will still be derived from the longer sequence.

 Even though our technique might give disparate answers to formulas
 $\alpha$ and $\beta$ which have different languages, the answer set
 for any fixed set of subject matters will have quite nice properties.
 \begin{observation} If $\Sigma$ is a set of propositional atoms, then $\{
 \alpha | L_\alpha = \Sigma \} \cap C_k(\sigma) \not\vdash \perp$.
 \end{observation}
 The agent's responses to a particular subject matter are
 guaranteed to be consistent by the query answering scheme.\vspace{.05in}

 \noindent {\bf Remark:} Our inference method corresponds to the {\it
 liberal} inference defined on a linearly prioritized sequence of formulas
 in \cite{kg96}. \cite{kg96} defines a {\it strict} notion of inference as
 well, which in our case would correspond to stopping the construction of
 $\Gamma_{\langle\sigma,k,\gamma\rangle}$ upon encountering the first
 formula that would make the set inconsistent.
 \subsection {Properties of $\vdash_k$}

 \begin{proposition}The following properties hold for the process of
 revision defined on sequences. \begin{itemize}
 \item Weak Inclusion: If
 $\gamma \not =\perp$ then $\sigma \ast \gamma \vdash_k \gamma$
 \end{itemize} The following additional properties hold {\em under the
 condition}\, that $L_{\alpha} = L_{\beta}$: \begin{itemize}
 \item Weak (or Cautious) Monotonicity:
 \[\frac{\sigma \vdash_k \alpha, \sigma \vdash_k \beta}{\sigma \ast \alpha
 \vdash_k \beta}\] \item Rational Monotonicity: \[\frac{\sigma \not\vdash_k
 \neg\alpha, \sigma \vdash_k \beta}{\sigma \ast \alpha \vdash_k \beta}\]
 \item Weak Cut: \[\frac{\sigma \ast \alpha \vdash_k \beta, \sigma \vdash_k
 \alpha}{\sigma \vdash_k \beta}\]
 \item Adjunction: \[\frac{\sigma \vdash_k
 \alpha, \sigma \vdash_k \beta}{\sigma \vdash_k \alpha \wedge \beta}\] \item
 Right Weakening: \[\frac{\sigma \vdash_k \alpha, \alpha \vdash
 \beta}{\sigma \vdash_k \beta}\]

 \end{itemize} \end{proposition} The condition $L_{\alpha} = L_{\beta}$ is
 important for the following reason.  As we have remarked, $[\![\sigma]\!]$
 may be, and usually is, inconsistent.  To get consistent answers to a query
 $\gamma$ we restrict ourselves to a certain portion of $[\![\sigma]\!]$ and
 this portion will depend on the language $L_{\gamma}$.  It is natural then
 that different $\gamma,\gamma'$ with different languages will evoke
 different subsets of $[\![\sigma]\!]$.  We cannot then require that these
 subsets will always produce coherent answers which obey our rules.
 However, in practice there will be {\em some}\, coherence.

 \noindent For
 just one example that Weak Monotonicity can fail in general unless
 $L_{\alpha}
 = L_{\beta}$: let $\sigma
 = [p, \neg p \wedge \neg q, q]$.  Now, $\forall k, \sigma \vdash_k p \wedge
 q$, but also, $\sigma \vdash_k \neg p$.  But clearly, $\sigma \ast (p
 \wedge q) \not\vdash_k \neg p$.  Similar examples can be constructed for
 the other rules mentioned above.  From a practical point of view, if two
 queries $\alpha, \beta$
 are asked on the {\em same}\, occasion, a smarter query answering
 procedure should  use $L
 = L_\alpha \cup L_\beta$ as the language for determining direct relevance.
 The answers thus generated will be compatible with each other.


 New formulas can block the derivation of formulas which were derivable
 before and so they provide a simple modeling for loss of belief in a
 proposition.
 After all, agents do not lose beliefs without a reason: to
 drop the belief that $\alpha$ is to {\it revise} by some information that
 changes our reasoning.  Still, it is possible in our model to lose
 $\alpha$ without acquiring $\neg\alpha$.  For example, consider the
 sequences $\sigma = p \wedge q$ and $\sigma \ast (\neg p \vee \neg q)$.
 The revised sequence no longer answers `yes` to $p$ but neither does it
 answer `yes` to $\neg p$. $(\neg p \vee \neg q)$ has undermined
 $\sigma = p \wedge q$ without actually making $\neg p$ derivable.
 \subsubsection{Equivalence}
 Given the notion of inference defined above, we say that belief sequences
 which yield the same answers to all queries are equivalent.  But
 equivalence is not always preserved under extensions.  Consider the
 following example: \noindent $\sigma_1 = [\neg p \wedge \neg q], \sigma_2 =
 [p, \neg p \wedge \neg q]$. $\sigma_1, \sigma_2$ are equivalent, but neither $\sigma_1$ nor
$\sigma_2$ implies
 $p$ as a conclusion.  However, revising by the formula $p \vee
 q$ yields $p$ as a conclusion for $\sigma_2$ though not for $\sigma_1$.
 Given this observation, we would like to define a strong
 notion of equivalence which is unaffected by revision.
\begin{definition}\begin{itemize}
\item
Sequences $\sigma_1, \sigma_2$ are equivalent if $C(\sigma_1) =
 C(\sigma_2)$.
\item Sequences $\sigma_1, \sigma_2$ are strongly
equivalent if $C(\sigma_1 \ast \alpha_1 \ast \ldots \alpha_n) =
C(\sigma_1 \ast \alpha_1 \ast \ldots \alpha_n)$ for all sequences of
revisions $\alpha_1, \ldots, \alpha_n$. \item $\sigma_1$ subsumes $\sigma_2$
 if $C(\sigma_2) \subseteq C(\sigma_1)$.
\end{itemize} \end{definition}
 We suspect that the notion of strong equivalence is testable via a
 single revision, i.e., if $\sigma$ and $\sigma'$ are not strongly
 equivalent, then there is a $\gamma$ such that $\sigma\ast\gamma$ and
 $\sigma '\ast\gamma$ are not equivalent.

 An obvious, related, question is, how can all sequences be trimmed or
 reduced to their shortest equivalent form?  The task of reducing sequences
 to their simplest equivalent form is most likely, computationally
 non-trivial.

 \subsubsection{Complexity of the Inference Procedure} In the methods
 defined above, there are two sources of complexity.  For any query $\gamma$
 the first one involves the calculation of the smallest language
 $L_{\gamma}$, which is a co-{\bf NP} complete problem \cite{ho99}, but
 only in the (relatively short) length of the individual formula $\gamma$
 and not in the size of the entire belief base $\sigma$.  The second one
 involves checking the consistency of the set $\Gamma_i$ at each step of the
 construction of the set $\Gamma$ of the maxiconsistent set related to the
 query formula $\gamma$.  However, for normal bases $\sigma$ it is likely to
 be the case that symbols which occur in formulas which are $k$-relevant to
 $\gamma$ are few in number.  In other words, there is a small language $L'$
 with $|L'| \ll |L|$ such that for any formula $\alpha \in \sigma$, with
 $\alpha$ $k$-relevant to $\gamma$, $L_{\alpha} \subseteq L'$. The
 consistency check then will be quite feasible.

 Our answering method first calculates the relevance relation exhaustively
 for the entire sequence.  Then the set $\Gamma$ is constructed formula by
 formula.  Since each stage of the construction involves a consistency
 check, the complexity of the procedure is polynomial with an {\bf NP}
 oracle, but only in $|L'|$ which will be small if we keep $k$ small.
 Moreover, in checking for $k$-derivability, the costs would be reduced
 sharply as most formulas in the sequence would not be $k$-relevant and the
 size of $\Gamma_{(\sigma,k,\gamma)}$ could be quite small.  This is indeed
 how we reason in practice.  While the logical problem of deriving
 consequences from our (possibly inconsistent) beliefs is daunting, the
 notion of relevance cuts down sharply on the effort involved.

\subsubsection{Comparison with $B$-structures}
 In \cite{cp99} a method for answering queries from $B$-structures is
 presented that allowed the answer $\top$, i.e., ({\it over-defined} or {\it
 inconsistent}).  In our method of inference, we ensure that
 $\Gamma_{\langle\sigma,k,\gamma\rangle}$ is always consistent, thus
 preventing an inconsistent answer.  The procedure of \cite{cp99} will
 (after suitable interpretation) agree with our method {\em when}\, the
 former gave 'yes' or 'no' answers.  The B-structure query answering method
 corresponds to $\vdash_0$, i.e., to $\Gamma_{<\sigma, 0, \gamma>}$
 constructed only with directly relevant formulas.
 \subsubsection{Conformance with AGM axioms}
 It is natural to ask how our inference conforms to AGM-like postulates.  We
 cannot apply the results in~\cite{kg96} that show that maxiconsistent
 inference on a belief sequence is rational; our inference procedure takes
 into account relevance relations that cause a reordering of our sequence,
 depending on the formula whose inference we are testing.  However,
 some AGM-like properties do hold.

 \noindent ($\sigma^{\ast}$ 1) $\sigma \ast \gamma$ is a belief sequence

 \noindent ($\sigma^{\ast}$ 2) $\gamma \in C_k(\sigma \ast \gamma)$ provided
 that $\gamma$ is consistent.

 \noindent ($\sigma^{\ast}$ 3) If $ \gamma \Leftrightarrow \beta$
 then
 $\forall \gamma, \sigma, \sigma \ast \gamma \vdash_k \gamma $ iff $\sigma
 \ast \beta \vdash_k \gamma$.

\noindent Note too, that there is no counterpart to the AGM expansion operation $\pdot$ in
 our model; revision just is concatenation and the content of the new epistemic
 state upon revision is given by the inference operation defined above.

 \subsubsection{Conformance with Lehmann Axioms}
 A comparison with the Lehmann axioms shows that our method of inference
 defined above does not conform to them.  The only one that the method
 conforms to is {\bf I2} which says that $\alpha \in [\sigma\circ\alpha]$.
 The reason is that in addition to the temporal
 ordering of the belief set, we have also imposed a relevance ordering.  The
 combination of these two orderings ensures that formulas whose inference
 was possible at one stage of the inference procedure could be blocked at
 later stages in the history of the agent not just by newer information but
 also by the ancillary phenomenon of more relevant beliefs moving to the
 front of the reorganized sequence.  Thus if we acquire a new belief which
 links $p,q$, then a query about $p$ will regard formulas involving $q$ as
 1-relevant.  For example, after China invades Tibet (and we find out about it)
 beliefs about China may move to the front of the relevance ordering, even
 in a query which is originally only about Tibet.  We believe ours is the
 first method of updating and revision which takes such phenomena into
 account.  Granted that some rather elegant properties of other formalisms are
 lost in our method, or, rather, hold only under conditions like $L_{\alpha}
 = L_{\beta}$.  However, this is a {\em necessary}\, consequence of the
 fact that we
 are handling sophisticated and delicate mechanisms used by actual agents to
 balance the need for coherenece with the limits on resources.
 \section{Conclusion}
 The method of inference (and belief revision) described is easily
 implemented via a finite representation of the sequence.  A full revision
 history is maintained and iterated revision is routine.  Explicit attention
 is paid to temporal ordering and relevance relations.  Our method extends
 intuitions found in the work on belief bases, sequences and inference
 relations based on prioritizations of beliefs and goes beyond by making
 clear the importance of a plausible notion of relevance to guide inference.
 The method presented is (for small $k$) resource conscious; in future work,
 we plan to implement this method and further investigate its properties.
\bibliographystyle{aaai}
\bibliography{newcgp}

\begin{thebibliography}{}

\bibitem[\protect\citeauthoryear{Alchourron, G\"ardernfors, \&
  Makinson}{1985}]{agm85}
Alchourron, C.; G\"ardernfors, P.; and Makinson, D.
\newblock 1985.
\newblock On the logic of theory change: Partial meet functions for contraction
  and revision.
\newblock {\em Journal of Symbolic Logic} 50:510--530.

\bibitem[\protect\citeauthoryear{Brewka}{1991}]{bre91}
Brewka, G.
\newblock 1991.
\newblock Belief revision in a framework for default reasoning.
\newblock In Fuhrmann, A., and Morreau, N., eds., {\em The Logic of Theory
  Change}, Lecture Notes in Artificial Intelligence, Number 465. Berlin:
  Springer.
\newblock  206--222.

\bibitem[\protect\citeauthoryear{Chopra \& Parikh}{1999}]{cp99}
Chopra, S., and Parikh, R.
\newblock 1999.
\newblock An inconsistency tolerant model for belief representation and belief
  revision.
\newblock In {\em Proceedings of the 16th International Joint Conference on
  Artificial Intelligence},  192--197.
\newblock Morgan-Kaufmann.

\bibitem[\protect\citeauthoryear{Doyle}{1979}]{do79}
Doyle, J.
\newblock 1979.
\newblock A truth maintenance system.
\newblock {\em Artificial Intelligence} 12:231--272.

\bibitem[\protect\citeauthoryear{Epstein}{1995}]{eps95}
Epstein, R.
\newblock 1995.
\newblock {\em The Semantical Foundations of Logic: Propositional Logics}.
\newblock Oxford University Press.

\bibitem[\protect\citeauthoryear{G\"ardenfors \& Makinson}{1988}]{gm88}
G\"ardenfors, P., and Makinson, D.
\newblock 1988.
\newblock Revisions of knowledge systems using epistemic entrenchment.
\newblock In Vardi, M., ed., {\em Proceedings of Theoretical Aspects of
  Reasoning about Knowledge},  83--96, 661--672.
\newblock Morgan Kaufmann.

\bibitem[\protect\citeauthoryear{G\"ardenfors \& Rott}{1995}]{gr95}
G\"ardenfors, P., and Rott, H.
\newblock 1995.
\newblock Belief revision.
\newblock In Gabbay, D.~M.; Hogger, C.~J.; and Robinson, J.~A., eds., {\em
  Handbook of Logic in Artificial Intelligence and Logic Programming, Volume
  IV: Epistemic and Temporal Reasoning}. Cambridge: Oxford University Press.
\newblock  35--132.

\bibitem[\protect\citeauthoryear{Georgatos}{1996}]{kg96}
Georgatos, K.
\newblock 1996.
\newblock Ordering-based representations of rational inference.
\newblock In Alferes, J.; Pereira, L.; and Orlowska, E., eds., {\em Logics in
  Artificial Intelligence (JELIA '96)}, Lecture Notes in Artificial
  Intelligence,  176--191.
\newblock Berlin: Springer-Verlag.

\bibitem[\protect\citeauthoryear{Georgatos}{1997}]{kg97}
Georgatos, K.
\newblock 1997.
\newblock Entrenchment relations: A uniform approach to nonmonotonicity.
\newblock In {\em Proceedings of the International Joint Conference on
  Qualitative and Quantitative Practical Reasoning (ESCQARU/FAPR 97)}, Lecture
  Notes in Computer Science, Number 1244,  282--297.
\newblock Berlin: Springer-Verlag.

\bibitem[\protect\citeauthoryear{Hansson}{1992}]{han92}
Hansson, S.~O.
\newblock 1992.
\newblock In defense of base contraction.
\newblock {\em Synthese} 91:239--245.

\bibitem[\protect\citeauthoryear{Herzig \& Rifi}{1999}]{ho99}
Herzig, A., and Rifi, O.
\newblock 1999.
\newblock Propositional belief base update and minimal change.
\newblock {\em Artificial Intelligence} 115(1):107--138.

\bibitem[\protect\citeauthoryear{Lehmann}{1995}]{leh95}
Lehmann, D.
\newblock 1995.
\newblock Belief revision, revised.
\newblock In {\em Proceedings of the Fourteenth International Joint Conference
  on Artificial Intelligence},  1534--1540.

\bibitem[\protect\citeauthoryear{Nebel}{1992}]{neb92}
Nebel, B.
\newblock 1992.
\newblock Syntax based approaches to belief revision.
\newblock In G\"ardenfors, P., ed., {\em Belief Revision}, Theoretical Computer
  Science. Cambridge: Cambridge University Press.

\bibitem[\protect\citeauthoryear{Parikh}{1999}]{par96}
Parikh, R.
\newblock 1999.
\newblock Beliefs, belief revision, and splitting languages.
\newblock In Moss, L.; Ginzburg, J.; and de~Rijke, M., eds., {\em Logic,
  Language, and Computation, Volume 2}, CSLI Lecture Notes No. 96. CSLI
  Publications.
\newblock  266--268.
\newblock Initially presented in {\it Preliminary Proceedings of Information
  Theoretic Approaches to Logic, Language, and Computation 1996}.

\bibitem[\protect\citeauthoryear{Rodrigues}{1997}]{rod97}
Rodrigues, O.~T.
\newblock 1997.
\newblock {\em A methodology for iterated information change}.
\newblock Ph.D. Dissertation, Imperial College, University of London.

\bibitem[\protect\citeauthoryear{Ryan}{1991}]{ry91}
Ryan, M.~D.
\newblock 1991.
\newblock Ordered theory presentations.
\newblock In {\em Proceedings of the 8th Amsterdam Colloquium}.

\bibitem[\protect\citeauthoryear{Wasserman}{1999}]{was99}
Wasserman, R.
\newblock 1999.
\newblock Resource bounded belief revision.
\newblock {\em Erkenntnis}.
\newblock To appear.

\end{thebibliography}
\end{document}